%% file: main.tex
\documentclass[conference]{IEEEtran}
\usepackage{cite}
\usepackage{hyperref}

\ifCLASSINFOpdf
   \usepackage[pdftex]{graphicx}
\else
  \usepackage[dvips]{graphicx}
\fi
\usepackage{algorithmic}
\usepackage{algorithm}
\usepackage{dblfloatfix}
\usepackage{placeins}

\begin{document}
\title{A Self-Driving Robot Using Deep Convolutional Neural Networks on Neuromorphic Hardware}

\author{\IEEEauthorblockN{Tiffany Hwu\IEEEauthorrefmark{1}\IEEEauthorrefmark{2},
Jacob Isbell\IEEEauthorrefmark{3},
Nicolas Oros\IEEEauthorrefmark{4}, 
and Jeffrey Krichmar\IEEEauthorrefmark{1}\IEEEauthorrefmark{5}}
\IEEEauthorblockA{\IEEEauthorrefmark{1}Department of Cognitive Sciences\\
University of California, Irvine\\
Irvine, California, USA, 92697}
\\
\IEEEauthorblockA{\IEEEauthorrefmark{2}Northrop Grumman \\
Redondo Beach, California, USA, 90278}
\\
\IEEEauthorblockA{\IEEEauthorrefmark{3}Department of Electrical and Computer Engineering\\
University of Maryland\\
College Park, Maryland, USA, 20742}
\\
\IEEEauthorblockA{\IEEEauthorrefmark{4}BrainChip LLC\\
Aliso Viejo, California, USA, 92656}
\\
\IEEEauthorblockA{\IEEEauthorrefmark{5}Department of Computer Sciences\\
University of California, Irvine\\
Irvine, California, USA, 92697\\
Email: jkrichma@uci.edu}
}


%Telephone: (800) 555--1212, Fax: (888) 555--1212}


\maketitle

\begin{abstract}
Neuromorphic computing is a promising solution for reducing the size, weight and power of mobile embedded systems. In this paper, we introduce a realization of such a system by creating the first closed-loop battery-powered communication system between an IBM TrueNorth NS1e and an autonomous Android-Based Robotics platform. Using this system, we constructed a dataset of path following behavior by manually driving the Android-Based robot along steep mountain trails and recording video frames from the camera mounted on the robot along with the corresponding motor commands. We used this dataset to train a deep convolutional neural network implemented on the TrueNorth NS1e. The NS1e, which was mounted on the robot and powered by the robot's battery, resulted in a self-driving robot that could successfully traverse a steep mountain path in real time. To our knowledge, this represents the first time the TrueNorth NS1e neuromorphic chip has been embedded on a mobile platform under closed-loop control. 
\end{abstract}


% \fi
\IEEEpeerreviewmaketitle

\section{Introduction}
\input{introduction.tex}
\section{Platforms}
\input{platforms.tex}
\section{Methods and Results}
\subsection{Data Collection}
\input{datacollection.tex}
\subsection{Eedn Framework}
\input{eednframework.tex}
\subsection{Data Pipeline}
\input{datapipeline.tex}
\subsection{Physical Connection of Platforms}
\input{physical.tex}
\subsection{Testing}
\input{testing.tex}

\section{Discussion}
\input{discussion.tex}

\section{Conclusion}
\input{conclusion.tex}

\section*{Acknowledgment}

The authors would like to thank Andrew Cassidy and Rodrigo Alvarez-Icaza of IBM for their support. This work was supported by the National Science Foundation Award number 1302125 and Northrop Grumman Aerospace Systems. We also would like to thank the Telluride Neuromorphic Cognition Engineering Workshop, The Institute of Neuromorphic Engineering, and their National Science Foundation, DoD and Industrial Sponsors. 



\FloatBarrier
\bibliographystyle{IEEEtran}
\bibliography{references}

\end{document}

%% file: introduction.tex
As the need for faster, more efficient computing continues to grow, the observed rate of improvement of computing speed shows signs of leveling off \cite{backus1978can}. In response, researchers have been looking for new strategies to increase computing power. Neuromorphic hardware is a promising direction for computing, taking a brain-inspired approach to achieve magnitudes lower power than traditional Von Neumann architectures \cite{mead1990neuromorphic,indiveri2011neuromorphic}. Mimicking the computational strategy of the brain, the hardware uses event-driven, massively parallel and distributed processing of information. As a result, the hardware has low size, weight, and power, making it ideal for mobile embedded systems. In exploring the advantages of neuromorphic hardware, it is important to consider how this approach might be used to solve our existing needs and applications. 

One such application is autonomous driving \cite{thrun2010toward}. In order for an autonomous mobile platform to perform effectively, it must be able to process large amounts of information simultaneously, extracting salient features from a stream of sensory data and making decisions about which motor actions to take \cite{levinson2011towards}. Particularly, the platform must be able to segment visual scenes into objects such as roads and pedestrians \cite{thrun2010toward}. Deep convolutional networks (CNNs) \cite{lecun1989backpropagation} have proven very effective for many tasks, including self-driving. For instance, Huval et al. used deep learning on a large dataset of highway driving to perform a variety of functions such as object and lane detection \cite{huval2015empirical}. Recently, Bojarski et al., showed that tasks such as lane detection do not need to be explicitly trained \cite{bojarski2016end}. In their DAVE-2 network, an end-to-end learning scheme was presented in which the network is simply trained to classify images from the car's cameras into steering commands learned from real human driving data. Intermediate tasks such as lane detection were automatically learned within the intermediate layers, saving the work of selecting these tasks by hand.

Such networks are suitable for running on neuromorphic hardware due to the large amount of parallel processing involved. In fact, many computer vision tasks have already been successfully transferred to the neuromorphic domain, such as handwritten digit recognition \cite{lee2016training} and scene segmentation \cite{cao2015spiking}. However, less work has been done embedding the neuromorphic hardware on mobile platforms. An example includes NENGO simulations embedded on SpiNNaker boards controlling mobile robots \cite{conradt2015trainable,galluppi2014event}. Addressing the challenges of physically connecting these components, as well as creating a data pipeline for communication between the platforms is an open issue, but worth pursuing given the small size, weight and power of neuromorphic hardware. 
% * <jkrichma@uci.edu> 2016-11-02T01:48:15.086Z:
%
% > An example includes NENGO simulations embedded on SpiNNaker boards controlling mobile robots (J. Conradt, F. Galluppi, and T. C. Stewart, "Trainable sensorimotor mapping in a neuromorphic robot," Robotics and Autonomous Systems, vol. 71, pp. 60-68, Sep 2015; F. Galluppi, C. Denk, M. C. Meiner, T. C. Stewart, L. A. Plana, C. Eliasmith, et al., "Event-based neural computing on an autonomous mobile platform," 2014 Ieee International Conference on Robotics and Automation (Icra), pp. 2862-2867, 2014.). 
%
% Add these references.
%
% ^.

At the Telluride Neuromorphic Cognition Workshop 2016, we embedded the the IBM TrueNorth NS1e \cite{merolla2014million} on the Android-Based Robotics platform \cite{oros2013smartphone} to create a self-driving robot that uses a deep CNN to travel autonomously along an outdoor mountain path. The result of our experiment is a robot that is able to use video frame data to steer along a road in real time with low-powered processing.
% * <jkrichma@uci.edu> 2016-11-02T03:12:27.966Z:
%
% > \cite{oros2013smartphone}
%
% incomplete reference.
%
% ^.

%% file: platforms.tex
\subsection{IBM TrueNorth}
\begin{figure}[!h]
 \centering
 \includegraphics[scale = .35]{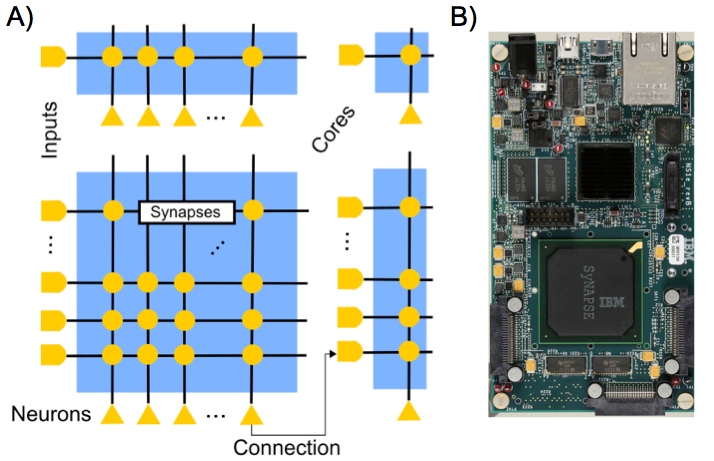}
 \caption{A) Core connectivity on the TrueNorth. Each neuron on a core connects to every other neuron on the core, and can connect to other cores through input lines. B) The IBM NS1e board. Adapted from \cite{esser2016convolutional}. }
 \label{trueNorth}
\end{figure}
The IBM TrueNorth (Figure \ref{trueNorth}) is a neuromorphic chip with a multicore array of programmable neurons. Within each core, there are 256 input lines connected to 256 neurons through a 256x256 synaptic crossbar array. Each neuron on a core is connected with every other neuron on the same core through the crossbar, and can communicate with neurons on other cores through their input lines. In our experiment, we used the IBM NS1e board, which contains 4096 cores, 1 million neurons, and 256 million synapses. An integrate-and-fire neuron model having 23 parameters was used, with trinary synaptic weights of -1, 0, and 1.  As the TrueNorth has been used to run many types of deep convolutional networks, and is able to be powered by an external battery, it served as ideal hardware for this task \cite{akopyan2016design} \cite{esser2016convolutional} .

\subsection{Android Based Robotics}
\begin{figure}[!h]
 \centering
 \includegraphics[scale = .9]{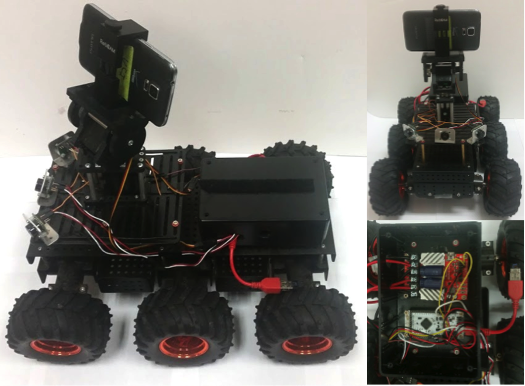}
 \caption{Left: Side view of CARLrado. A pan and tilt unit supports the Samsung Galaxy S5 smartphone, which is mounted on a Dagu Wild Thumper chassis. A plastic enclosure holds the IOIO-OTG microcontroller and RoboClaw motor controller. A velcro strip on top of the housing can attach any other small components. Top Right: Front view of CARLrado. Three front-facing sonars can detect obstacles. Bottom Right: Close-up of IOIO-OTG and motor controller.}
 \label{abr}
\end{figure}
The Android-Based Robotics platform (Figure \ref{abr}) was created at the University of California, Irvine, using entirely off-the-shelf commodity parts and controlled by an Android phone \cite{oros2013smartphone}.  The robot used in the present experiment, the CARLorado, was constructed from a Dagu Wild-Thumper All-Terrain chassis that could easily travel through difficult outdoor terrain. A IOIO-OTG microcontroller (SparkFun Electronics) communicated through a Bluetooth connection with the Android phone (Samsung Galaxy S5). The phone provided extra sensors such as a built-in accelerometer, gyroscope, compass, and global positioning system (GPS). The IOIO-OTG controlled a pan and tilt unit that held the phone, a motor controller for the robot wheels, and ultrasonic sensors for detecting obstacles. Instructions for building the robot can be found at: \url{http://www.socsci.uci.edu/~ jkrichma/ABR/}. A differential steering technique was used, moving the left and right sides of the robot at different speeds for turning. The modularity of the platform made it easy to add extra units such as the IBM TrueNorth.
% * <jkrichma@uci.edu> 2016-11-02T03:20:18.676Z:
%
% > http://www.socsci.uci.edu/~jkrichma/ABR/
%
% The ~ symbol is not showing up in the PDF.
%
% ^.

Software for controlling the robot was written in Java using Android Studio. With various support libraries for the IOIO-OTG, open-source libraries for computer vision such as OpenCV, and sample Android-Based Robotics code (https://github.com/UCI-ABR), it was straightfoward to develop intelligent controls.

%% file: datacollection.tex
\begin{figure}[!h]
 \centering
 \includegraphics[scale = .40]{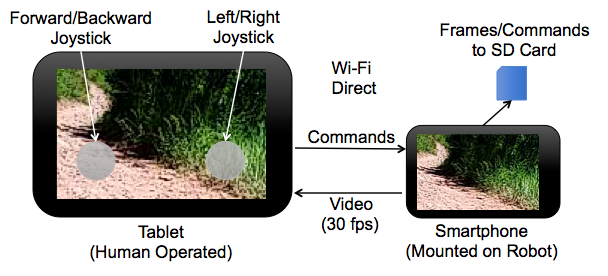}
 \caption{Data collection setup. Video from the smartphone mounted on the robot was sent to the tablet through a Wi-Fi direct connection.  A human operator used two joysticks on the touchscreen of the tablet to issue motor commands, which were sent to the phone through the same connection. Video frames and commands were saved to the SD card on the smartphone.}
 \label{dataCollection}
\end{figure}
\begin{figure}[h]
 \centering
 \includegraphics[scale = .35]{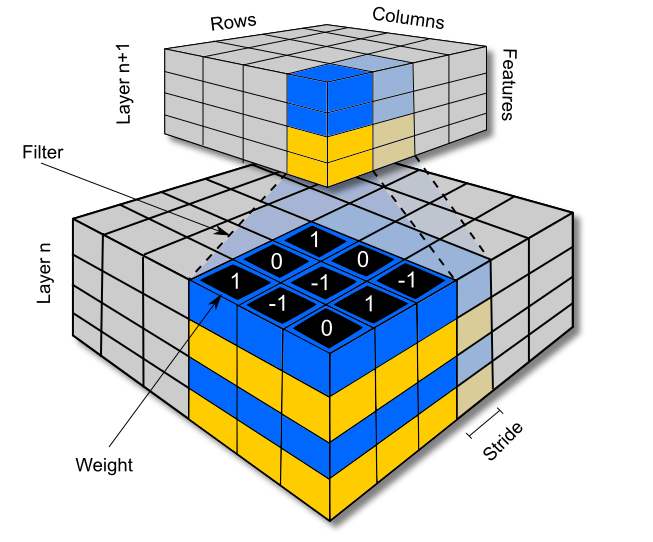}
 \caption{Convolution of layers in a CNN on TrueNorth. Neurons in each layer are arranged in three dimensions, which can be convolved using a filter of weights. Convolution occurs among the first two dimensions, and the third dimension represents different features. This allows the convolution to be divided along the feature dimension into groups (indicated by blue and yellow colors) that can be computed separately on different cores. Adapted from \cite{esser2016convolutional}. }
 \label{conv}
\end{figure}

First, we created datasets of first-person video footage of the robot and motor commands issued to the robot as it was manually driven along a mountain trail in Telluride, Colorado (Figures \ref{classifier} and \ref{mountainRoad} top). This was done by creating an app in Android Studio that was run on both a Samsung Galaxy S5 smartphone and a Samsung Nexus 7 tablet (Figure \ref{dataCollection}). The smartphone was mounted on the pan and tilt unit of the robot with the camera facing ahead. JPEG images captured by the camera of the smartphone were saved to an SD card at 30 frames per second. The JPEGs had a resolution of 176 by 144 pixels. Through a Wi-Fi direct connection, the video frame data was streamed from the phone to a handheld tablet that controlled the robot. The tablet displayed controls for moving the robot forward and steering the robot left and right. These commands from the tablet (left, right, forward) were streamed to the smartphone via the Wi-Fi direct connection and saved on the smartphone as a text file. A total of 4 datasets were recorded on the same mountain trail, with each dataset recording a round trip of .5 km up and down a single trail segment. To account for different lighting conditions, we spread the recordings across two separate days, and on each day we performed one recording in the morning and one in the afternoon. In total we collected approximately 30 minutes of driving data. By matching the time stamps of motor commands to video images, we were able to determine which commands corresponded to which images. Images that were not associated with a left, right, or forward movement such as stopping were excluded. Due to lack of time, only the first day of data collection was used in actual training.

%% file: eednframework.tex
\begin{figure}[h]
 \centering
 \includegraphics[scale = .40]{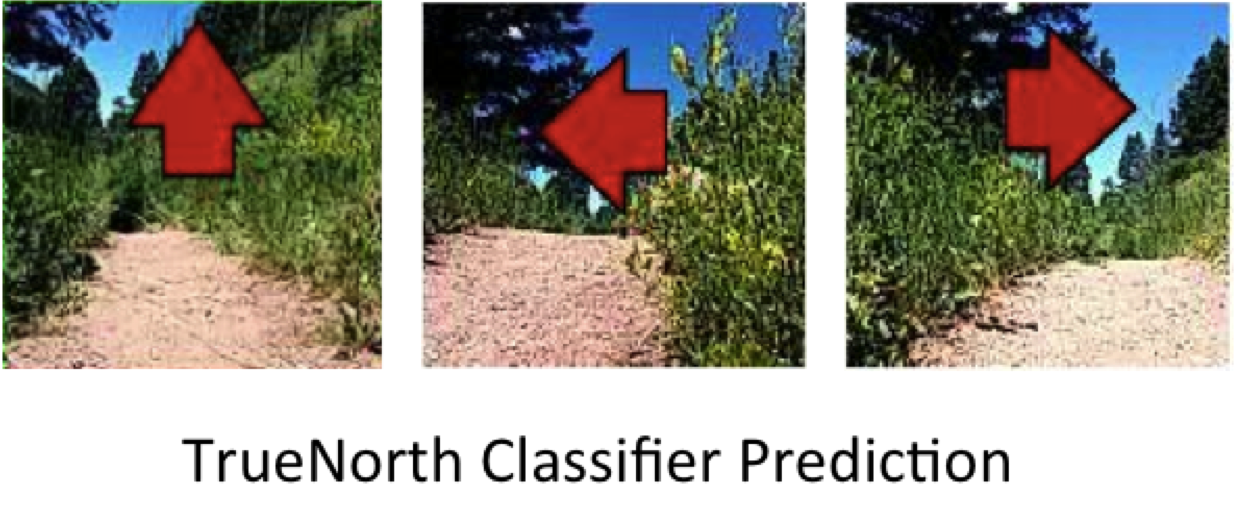}
 \caption{The CNN classified images into three classes of motor output: turning left, moving forward, and turning right. Accuracy of training was above 90 percent.}
 \label{classifier}
% * <jkrichma@uci.edu> 2016-10-13T04:57:45.808Z:
%
% >  \caption{The CNN classified images into three classes of motor output: turning left, moving forward, and turning right.}
%
% You need to explain how the driver commands were tied to the images in the training set. And it would be good to add more about how the CNN was trained. How many generations, how long did it take, what was the error? Do we have any of that information?
%
% ^ <hwut@uci.edu> 2016-10-25T22:26:25.853Z.
\end{figure}

\begin{figure}[h]
 \centering
 \includegraphics[scale = .25]{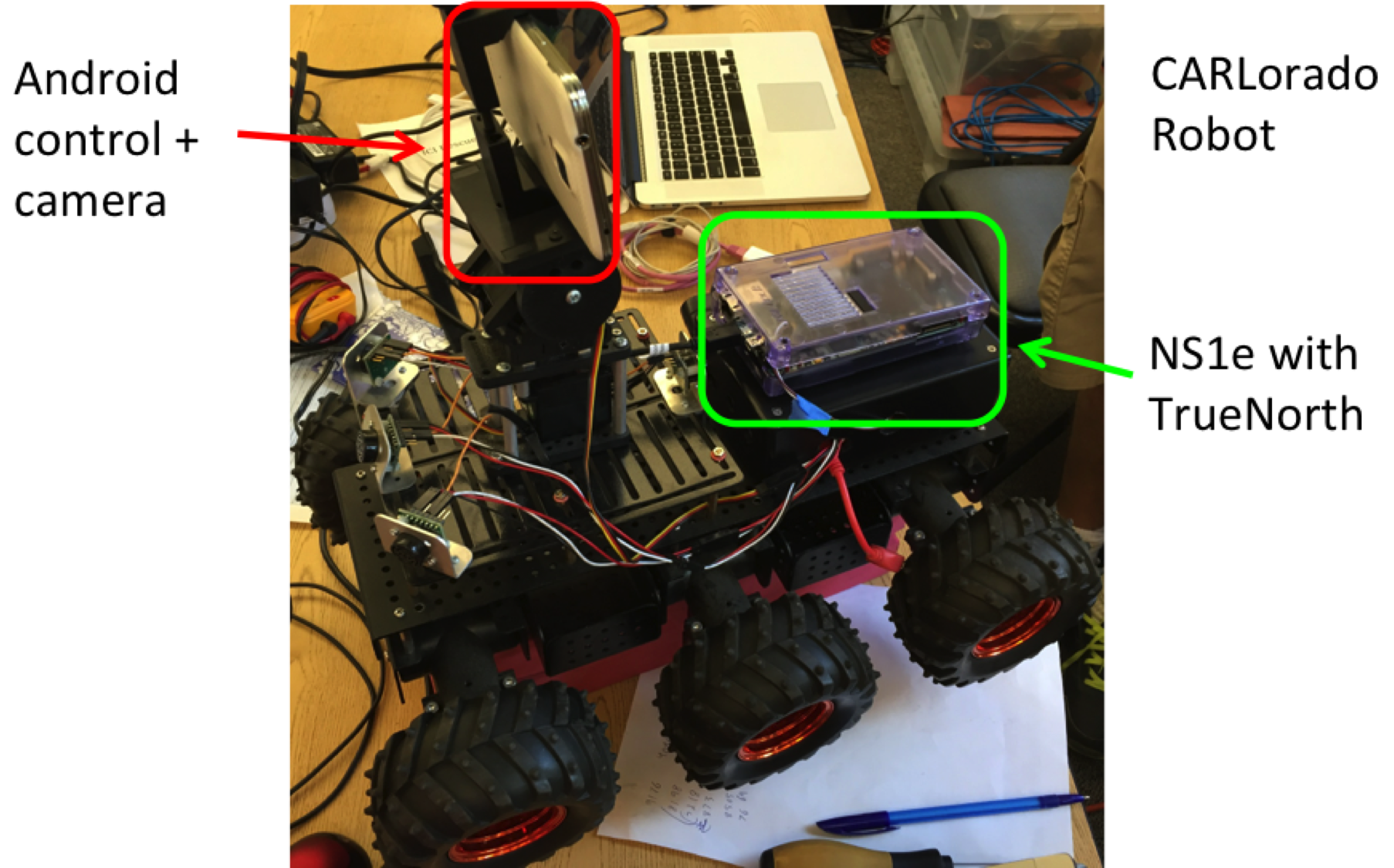}
 \caption{Physical connection of TrueNorth NS1e and CARLorado. The NS1e is attached to the top of the housing of the electronics housing using velcro. The NS1e is powered by running connections from the motor controller within the housing. The motor controller itself is powered by a Ni-MH battery attached to the bottom of the robot chassis.}
 \label{abrTrueNorth}
\end{figure}

\begin{figure*}[h]
 \centering
 \includegraphics[scale = .50]{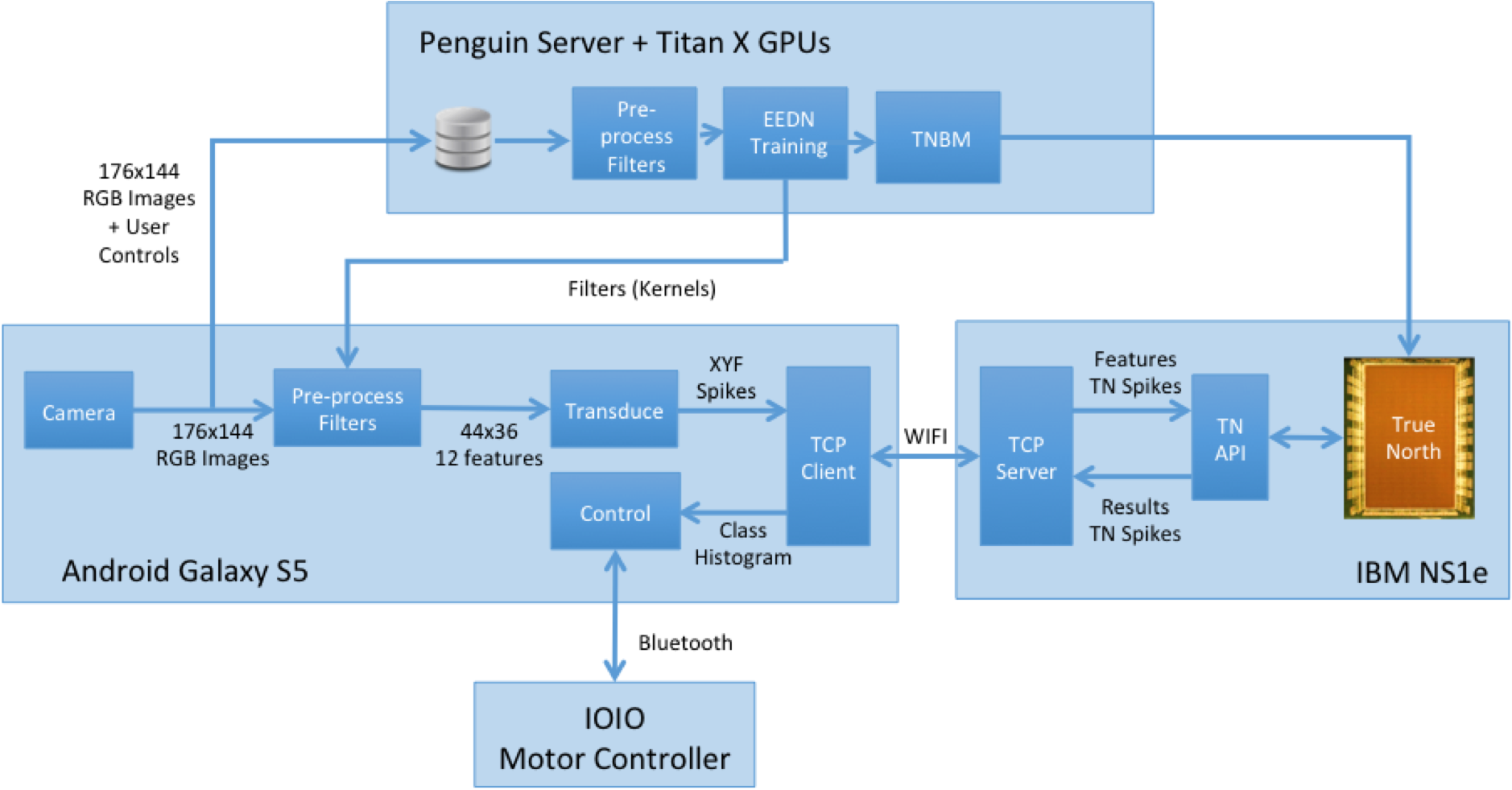}
 \caption{Data pipeline for running CNN. Training is done separately using the MatConvNet package using Titan X GPUs. A Wi-Fi connection between the Android Galaxy S5 and IBM NS1e transmit spiking data back and forth.}
 \label{pipeline}
\end{figure*}
\begin{figure}[!h]
 \centering
 \includegraphics[scale = .60]{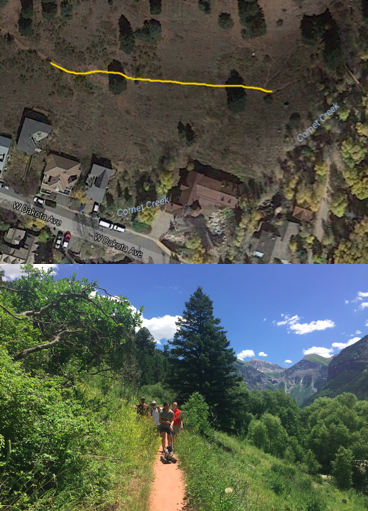}
 \caption{Mountain trail in Telluride, Colorado. Above: Google Satellite image of trail (highlighted) Imagery \copyright 2016 Google. Below: Testing CNN performance.}
 \label{mountainRoad}
\end{figure}
We  used the dataset to train a deep convolutional neural network using an Energy-Efficient Deep Neuromorphic Network (EEDN), a network that is structured to run efficiently on the TrueNorth\cite{esser2016convolutional}. In summary, a traditional CNN is transferred to the neuromorphic domain by connecting the neurons on the TrueNorth with the same connectivity as the original CNN. Input values to the original CNN are translated into input firing patterns on EEDN, and the resulting firing rates of each neuron correspond to the values seen in the original CNN. To distribute a convolutional operation among cores of the TrueNorth, the layers are divided along the feature dimension into groups (Figure \ref{conv}). When a neuron targets multiple core inputs, exact duplicates of the neuron and synaptic weights are created, either on the same core or a different core. The response of each neuron is the binary thresholded sum of synaptic input, in which the trinary weight values are determined by different combinations of two input lines. A more complete explanation of the EEDN flow and structure of the convolutional network (1 chip version) can be found in \cite{esser2016convolutional}.

The video frames were preprocessed by down-sampling them to a resolution of 44 by 36 pixels and separating them into red, green, and blue channels. The output is a single layer of three neuron populations, corresponding to three classes of turning left, going straight, or turning right, as seen in Figure \ref{classifier}. 

Using the MatConvNet package, a Matlab toolbox for implementing convolutional neural networks, the network was trained to classify images into motor commands. For instance, if the image showed the road to be more towards the left of center, the CNN would learn the human-trained command of steering to the left. To test accuracy, the dataset was split into train and test sets by using every fifth frame as a test frame (in total 20 percent of the dataset). We achieved an accuracy of over 90 percent, which took 10K iterations and a few hours to train.  Training was performed
% * <jkrichma@uci.edu> 2016-11-02T03:28:00.691Z:
%
% > MatConvNet package
%
% Define and explain what this is.
%
% ^.
separately from the TrueNorth chip, producing trinary synaptic weights (-1,0,1) that could be used interchangeably in a traditional CNN or EEDN.
% * <jkrichma@uci.edu> 2016-10-13T04:35:00.520Z:
%
% > Due to lack of time, only the first day of data collection was used in actual training.
%
% What do you mean? 4 datasets were collected. You didn't mention multiple days. Do you mean that only one dataset out of 4 was used?
%
% ^ <hwut@uci.edu> 2016-10-13T17:38:15.576Z.

%% file: datapipeline.tex
With the methods used in \cite{esser2016convolutional}, the weights of the network were transferred to the TrueNorth NS1e. The CNN was able to run on the TrueNorth by feeding input from the camera on the Android Galaxy S5 to the TrueNorth using a TCP/IP connection. In order to achieve this, the phone had to replicate the preprocessing used when training the network. The preprocessing on the phone was achieved by using the Android OpenCV scaling function to downsample the images. Then, the images were separated into red, green, and blue channels. Next, the filter kernels from the first layer of the CNN were pulled from the EEDN training output and applied to the image using a 2D convolution function from the Android OpenCV library. The result of the convolution was thresholded into binary spiking format, such that any neuron with an activity greater than zero was set to spike.  The spiking input to the TrueNorth was sent in XYF format, where X, Y, and F are the three dimensions to describe the identity of a spiking neuron within a layer. At each tick of the TrueNorth NS1e, a frame was fed into the input layer by sending the XYF coordinates of all neurons that spiked for that frame. A detailed diagram of the pipeline is found in Figure \ref{pipeline}. Output from the TrueNorth NS1e was sent back to the smartphone through the TCP/IP connection in the form of a class histogram, which indicated the firing activity of the output neurons. The smartphone could then calculate which output neuron was the most active and issue the corresponding motor command to the robot.
% * <jkrichma@uci.edu> 2016-10-13T04:37:41.485Z:
%
% >  Spiking input to the TrueNorth was sent in XYF format, where X and Y values determine the coordinates of an input neuron within the filter, and F determines the input firing rate. A detailed diagram of the pipeline is found in Figure \ref{pipeline}.
%
% Missing a step. How do you go from the filter kernels to spiking input?
%
% ^ <hwut@uci.edu> 2016-10-13T17:37:14.517Z:
%
% We may need to ask Andrew or Rodrigo for some clarification. The original IBM paper says that there is one classification done per tick,  but I don't remember if that's how we did it.
%
% ^ <hwut@uci.edu> 2016-10-25T22:30:24.645Z.

%% file: physical.tex
The TrueNorth was powered by connecting the robot's battery terminals from the motor controller to a two-pin battery connection on the NS1e board. It was then secured with velcro to the top of the housing for the IOIO and motor controller. A picture of the setup is seen in Figure \ref{abrTrueNorth}. The robot, microcontroller, motor controller, servos, and NS1e were powered by a single Duratrax	NiMH Onyx 7.2V 5000mAh battery.

%% file: testing.tex
With this wireless, battery-powered setup, the trained CNN was able to successfully drive the robot on the mountain trail (Figure \ref{mountainRoad}).  A wireless hotspot was necessary to create a TCP connection between the TrueNorth NS1e and the Android phone. We placed the robot on the same section of the trail used for training. The robot steered according to the class histograms received from the TrueNorth output, which provided total firing counts for each of the three output neuron populations. Steering was done by using the histogram to determine which output population fired the most, and steering in that direction. As a result, the robot stayed near the center of the trail, steering away from green brush on both sides of the trail. At some points, the robot did travel off the trail and needed to be manually redirected back towards the center of the trail. The robot drove approximately .5 km uphill, and the returned .5 km downhill with minimal intervention. It should be noted that there was a steep dropoff on the south side of the trail. Therefore, extra care was taken to make sure the robot did not tumble down the mountainside. A video of the path following performance can be seen at \url{https://www.youtube.com/watch?v=CsZah2hydeY}.

%% file: discussion.tex
To the best of our knowledge, the present setup represents the first time the TrueNorth NS1e has been embedded on a mobile platform under closed loop control. It demonstrated that a low power neuromorphic chip could communicate with a smartphone in an autonomous system. Furthermore, it showed that a CNN using the EEDN framework was sufficient to achieve a self-driving application. Furthermore, this complete system ran in real-time and was powered by a single off-the-shelf hobby grade battery, demonstrating the power efficiency of the TrueNorth NS1e. 

An expansion of this work would require better quantification of the robot's performance. This could be achieved by tracking the number of times the robot had to be manually redirected, or comparing the CNN classifier accuracy on the training set of images versus the classifier accuracy on the actual images captured in realtime.  Increasing the amount of training data would likely increase the classifier accuracy, since only 15 minutes of data were used for the training as compared to other self-driving CNNs \cite{huval2015empirical,bojarski2016end}, which have used several days or even weeks of training. Our success was due in part to the simplicity of the landscape, with an obvious red hue to the dirt road and bold green hue for the bordering areas. It would therefore be useful to test the network in more complex settings. Additionally, while the main purpose of the project was to demonstrate a practical integration of neuromorphic and non-neuromorphic hardware, it would also be useful to calculate the power savings of running the CNN computations on neuromorphic hardware instead of directly on the smartphone.

%% file: conclusion.tex
In this trailblazing study, we have demonstrated a novel closed-loop system between a robotic platform and a neuromorphic chip, operating in a rugged outdoor environment. We have shown the advantages of integrating neuromorphic hardware with popular machine learning methods such as deep convolutional neural networks. We have shown that neuromorphic hardware can be integrated with smartphone technology and off the shelf components resulting in a complete autonomous system. The present setup is one of the first demonstrations of using neuromorphic hardware in an autonomous, embedded system.% * <jkrichma@uci.edu> 2016-11-02T03:49:09.277Z:
%
% > groundbreaking
%
% Better word anyone?
%
% ^.
%
% > Through our project completed at the Telluride Neuromorphic Cognition Workshop, we have demonstrated a novel closed-loop system between a robotic platform and a neuromorphic chip, operating in a rugged outdoor environment. We have shown the advantages of integrating neuromorphic hardware with popular machine learning methods such as deep convolutional neural networks. We have shown that neuromorphic hardware can be integrated with smartphone technology and off the shelf components resulting in a complete autonomous system. The present setup is one of the first demonstrations of using neuromorphic hardware in an autonomous, embedded system.
%
% Should we delete this paragraph, as it is now stated at the beginning of the discussion?
%
% ^ <hwut@uci.edu> 2016-10-14T19:34:57.094Z.

%% file: main.bbl
\begin{thebibliography}{10}
\providecommand{\url}[1]{#1}
\csname url@samestyle\endcsname
\providecommand{\newblock}{\relax}
\providecommand{\bibinfo}[2]{#2}
\providecommand{\BIBentrySTDinterwordspacing}{\spaceskip=0pt\relax}
\providecommand{\BIBentryALTinterwordstretchfactor}{4}
\providecommand{\BIBentryALTinterwordspacing}{\spaceskip=\fontdimen2\font plus
\BIBentryALTinterwordstretchfactor\fontdimen3\font minus
  \fontdimen4\font\relax}
\providecommand{\BIBforeignlanguage}[2]{{%
\expandafter\ifx\csname l@#1\endcsname\relax
\typeout{** WARNING: IEEEtran.bst: No hyphenation pattern has been}%
\typeout{** loaded for the language `#1'. Using the pattern for}%
\typeout{** the default language instead.}%
\else
\language=\csname l@#1\endcsname
\fi
#2}}
\providecommand{\BIBdecl}{\relax}
\BIBdecl

\bibitem{backus1978can}
J.~Backus, ``Can programming be liberated from the von neumann style?: a
  functional style and its algebra of programs,'' \emph{Communications of the
  ACM}, vol.~21, no.~8, pp. 613--641, 1978.

\bibitem{mead1990neuromorphic}
C.~Mead, ``Neuromorphic electronic systems,'' \emph{Proceedings of the IEEE},
  vol.~78, no.~10, pp. 1629--1636, 1990.

\bibitem{indiveri2011neuromorphic}
G.~Indiveri, B.~Linares-Barranco, T.~J. Hamilton, A.~Van~Schaik,
  R.~Etienne-Cummings, T.~Delbruck, S.-C. Liu, P.~Dudek, P.~H{\"a}fliger,
  S.~Renaud \emph{et~al.}, ``Neuromorphic silicon neuron circuits,''
  \emph{Frontiers in neuroscience}, vol.~5, p.~73, 2011.

\bibitem{thrun2010toward}
S.~Thrun, ``Toward robotic cars,'' \emph{Communications of the ACM}, vol.~53,
  no.~4, pp. 99--106, 2010.

\bibitem{levinson2011towards}
J.~Levinson, J.~Askeland, J.~Becker, J.~Dolson, D.~Held, S.~Kammel, J.~Z.
  Kolter, D.~Langer, O.~Pink, V.~Pratt \emph{et~al.}, ``Towards fully
  autonomous driving: Systems and algorithms,'' in \emph{Intelligent Vehicles
  Symposium (IV), 2011 IEEE}.\hskip 1em plus 0.5em minus 0.4em\relax IEEE,
  2011, pp. 163--168.

\bibitem{lecun1989backpropagation}
Y.~LeCun, B.~Boser, J.~S. Denker, D.~Henderson, R.~E. Howard, W.~Hubbard, and
  L.~D. Jackel, ``Backpropagation applied to handwritten zip code
  recognition,'' \emph{Neural computation}, vol.~1, no.~4, pp. 541--551, 1989.

\bibitem{huval2015empirical}
B.~Huval, T.~Wang, S.~Tandon, J.~Kiske, W.~Song, J.~Pazhayampallil,
  M.~Andriluka, P.~Rajpurkar, T.~Migimatsu, R.~Cheng-Yue \emph{et~al.}, ``An
  empirical evaluation of deep learning on highway driving,'' \emph{arXiv
  preprint arXiv:1504.01716}, 2015.

\bibitem{bojarski2016end}
M.~Bojarski, D.~Del~Testa, D.~Dworakowski, B.~Firner, B.~Flepp, P.~Goyal, L.~D.
  Jackel, M.~Monfort, U.~Muller, J.~Zhang \emph{et~al.}, ``End to end learning
  for self-driving cars,'' \emph{arXiv preprint arXiv:1604.07316}, 2016.

\bibitem{lee2016training}
J.~H. Lee, T.~Delbruck, and M.~Pfeiffer, ``Training deep spiking neural
  networks using backpropagation,'' \emph{arXiv preprint arXiv:1608.08782},
  2016.

\bibitem{cao2015spiking}
Y.~Cao, Y.~Chen, and D.~Khosla, ``Spiking deep convolutional neural networks
  for energy-efficient object recognition,'' \emph{International Journal of
  Computer Vision}, vol. 113, no.~1, pp. 54--66, 2015.

\bibitem{conradt2015trainable}
J.~Conradt, F.~Galluppi, and T.~C. Stewart, ``Trainable sensorimotor mapping in
  a neuromorphic robot,'' \emph{Robotics and Autonomous Systems}, vol.~71, pp.
  60--68, 2015.

\bibitem{galluppi2014event}
F.~Galluppi, C.~Denk, M.~C. Meiner, T.~C. Stewart, L.~A. Plana, C.~Eliasmith,
  S.~Furber, and J.~Conradt, ``Event-based neural computing on an autonomous
  mobile platform,'' in \emph{2014 IEEE International Conference on Robotics
  and Automation (ICRA)}.\hskip 1em plus 0.5em minus 0.4em\relax IEEE, 2014,
  pp. 2862--2867.

\bibitem{merolla2014million}
P.~A. Merolla, J.~V. Arthur, R.~Alvarez-Icaza, A.~S. Cassidy, J.~Sawada,
  F.~Akopyan, B.~L. Jackson, N.~Imam, C.~Guo, Y.~Nakamura \emph{et~al.}, ``A
  million spiking-neuron integrated circuit with a scalable communication
  network and interface,'' \emph{Science}, vol. 345, no. 6197, pp. 668--673,
  2014.

\bibitem{oros2013smartphone}
N.~Oros and J.~L. Krichmar, ``Smartphone based robotics: Powerful, flexible and
  inexpensive robots for hobbyists, educators, students and researchers,''
  Center for Embedded Computer Systems, University of California, Irvine,
  Irvine, California, Tech. Rep. 13-16, 2013.

\bibitem{esser2016convolutional}
S.~K. Esser, P.~A. Merolla, J.~V. Arthur, A.~S. Cassidy, R.~Appuswamy,
  A.~Andreopoulos, D.~J. Berg, J.~L. McKinstry, T.~Melano, D.~R. Barch,
  C.~di~Nolfo, P.~Datta, A.~Amir, B.~Taba, M.~D. Flickner, and D.~S. Modha,
  ``Convolutional networks for fast, energy-efficient neuromorphic computing,''
  \emph{Proceedings of the National Academy of Sciences}, vol. 113, no.~41, pp.
  11\,441--11\,446, 2016.

\bibitem{akopyan2016design}
F.~Akopyan, ``Design and tool flow of ibm's truenorth: an ultra-low power
  programmable neurosynaptic chip with 1 million neurons,'' in
  \emph{Proceedings of the 2016 on International Symposium on Physical
  Design}.\hskip 1em plus 0.5em minus 0.4em\relax ACM, 2016, pp. 59--60.

\end{thebibliography}
